\documentclass{his04}

\usepackage{graphicx}

\begin{document}

\title{Cooperative Game Theory within Multi-Agent Systems for Systems Scheduling}
\date{}

\author{Derek Messie and Jae C. Oh}
\institute{
Department of Electrical Engineering and Computer Science\\
Syracuse University, Syracuse, NY 13244, USA\\
dsmessie@syr.edu, jcoh@ecs.syr.edu
}

\maketitle
\bibliographystyle{his04}

\noindent
\begin{center}
\begin{large}
\textbf{Abstract}  
\end{large}
\end{center}

\textit{Research concerning organization and coordination within multi-agent systems continues to draw from a variety of architectures and methodologies.  The work presented in this paper combines techniques from game theory and multi-agent systems to produce self-organizing, polymorphic, lightweight, embedded agents for systems scheduling within a large-scale real-time systems environment.  Results show how this approach is used to experimentally produce optimum real-time scheduling through the emergent behavior of thousands of agents.  These results are obtained using a SWARM simulation of systems scheduling within a High Energy Physics experiment consisting of 2500 digital signal processors.}

\begin{large}
\vspace*{.2in}
\noindent
\textbf{1.  Introduction}\\
\end{large}
\vspace*{-.1in}

Game theory has been used in a wide range of problems requiring coordination in large-scale complex systems \cite{cc:coopmultisys98}\cite{jc:powertransplan98}\cite{pd:computernetworkapps91}\cite{kp:qosprovision96}.  This paper describes a hybrid-intelligent, self-organizing, multi-agent systems approach to computer systems scheduling based on game theory.  The design is implemented on RTES/BTeV, a large-scale, real-time data acquisition system for a High Energy Physics (HEP) particle accelerator.  

Multiple layers of very lightweight agents (VLAs) are embedded within 2500 Digital Signal Processors (DSPs) to handle fault mitigation across the system.  One of the primary challenges is to determine the frequency at which VLAs should perform specific monitoring and mitigation tasks.  Results show how self-organizing VLAs within individual systems schedulers are used to experimentally find the optimum rate at which these fault mitigation and monitoring tasks should occur.  SWARM multi-agent simulation software is used to model the RTES/BTeV environment.

The paper is divided into four sections.  First, some background on the BTeV experiment and the RTES collaboration is provided, along with some details on VLAs embedded within Level 1 of the RTES/BTeV environment.  Current challenges and other motivating factors are also described.  The next section details the model for self-organizing VLAs within systems schedulers implemented on each of the 2500 DSPs.  This consists of a model overview and specifics on the self-organizing approach based on cooperative game theory.  The next section evaluates the results of a SWARM simulation of the RTES/BTeV environment that implements the self-organizing approach.  Finally, next steps are outlined, followed by a conclusion.

\begin{large}
\vspace*{.2in}
\noindent
\textbf{2.  Background and Motivation}\\
\vspace*{-.1in}

\noindent
\textbf{2.1  RTES/BTeV}
\vspace*{-.1in}
\vspace*{.2in}
\end{large}

BTeV is a proposed particle accelerator-based HEP experiment currently under development at Fermi National Accelerator Laboratory.  The goal is to study charge-parity violation, mixing, and rare decays of particles known as beauty and charm hadrons, in order to learn more about matter-antimatter asymmetries that exist in the universe today \cite{sk:pixeldetect02}.

\begin{figure*}[t]
\centering
\includegraphics[width=150mm]{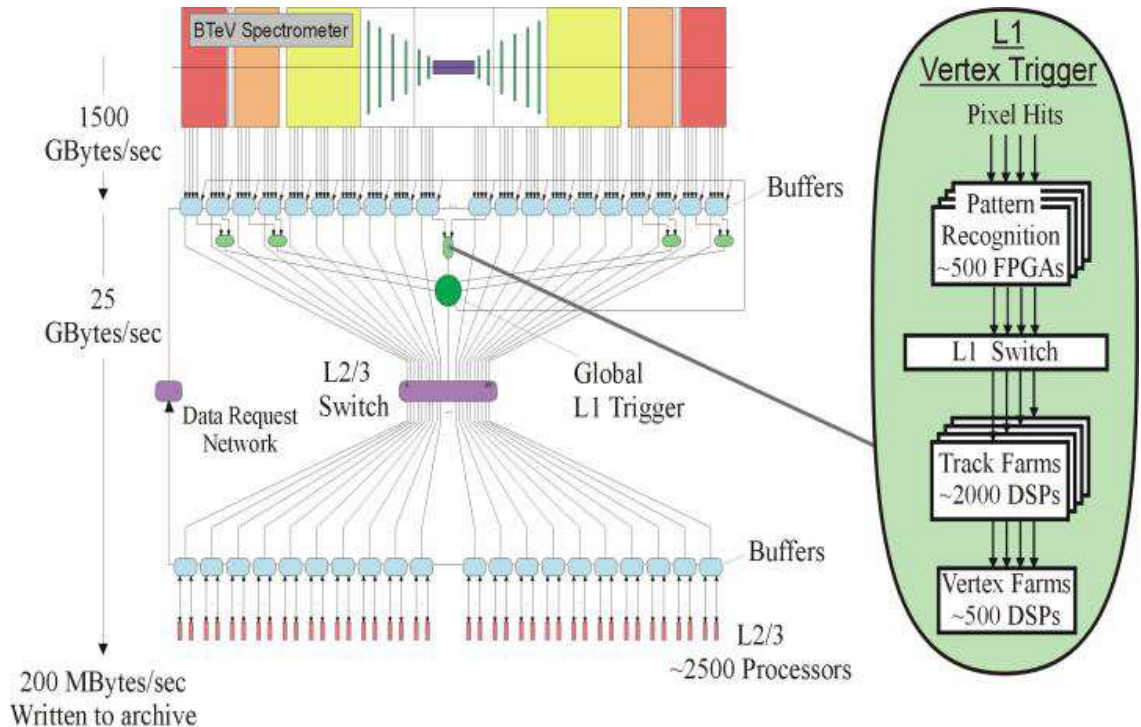}
\vspace*{-.7in}
\caption{The BTeV triggering and data acquisition system showing (left side) detector, buffer memories, L1, L2, L3 clusters and their interconnects and (right side) a magnified figure of the L1 Vertex trigger.}
\label{fig:btevtrigdata}
\end{figure*}

The experiment uses approximately 30 planar silicon pixel detectors that are connected to specialized field-programmable gate arrays (FPGAs).  The FPGAs are connected to approximately 2500 digital signal processors (DSPs) that filter incoming data at the extremely high rate of approximately 1.5 Terabytes per second from a total of 20x10$^{6}$ data channels.  A three tier hierarchical trigger architecture will be used to handle this high rate \cite{sk:pixeldetect02}.  An overview of the BTeV triggering and data acquisition system is shown in Figure \ref{fig:btevtrigdata}, including a magnified view of the L1 Vertex Trigger responsible for Level 1 filtering consisting of 2500 Worker nodes (2000 Track Farms and 500 Vertex Farms).

There are many Worker level tasks that the Farmlet VLA (FVLA) is responsible for monitoring.  A list of some of the tasks is shown in Figure \ref{fig:faultScenario}.  A traditional  hierarchical approach would assign one (or more) distinct DSPs the role of the FVLA, with the responsibility of monitoring the state of other Worker DSPs on the node.  However, this leaves the system with only very few possible points of failure before critical tasks are left unattended.

Another approach would be to assign a single DSP (or more) to each and every Worker DSP, to act as the FVLA.  However, since 2500 Worker DSPs are projected, this would prove very expensive and may still not fully protect all DSPs given even a low number of system failures.

The events that pass the full set of physics algorithm filters occur very infrequently, and the cost of operating this environment is high.  The extremely large streams of data resulting from the BTeV environment must be processed real-time with highly resilient adaptive fault tolerant systems.

\begin{large}
\vspace*{.2in}
\noindent
\vspace*{-.1in}
\textbf{2.2.  Very Lightweight Agents (VLAs)}
\vspace*{.2in}
\end{large}
\begin{figure*}[t]
\begin{center}
\begin{small}
\begin{tabular}{|p{.2in}|p{2.8in}|p{2.8in}|}

\hline
\textbf{ID} &\textbf{Description} &\textbf{Possible Causes}\\

\hline
e1 
& DSP over time budget on crossing processing. 
& Crossing was too complex to complete and developer was not careful to give up in time.\\
\hline
e2
& PA is stuck in a loop (within software timer control).
& Improper error handling caused the program to get stuck in an infinite loop.\\
\hline
e3
& DSP application framework is stuck in a loop (outside of software timer control).
& Logic error in code that manipulates the board's communication facilities.\\
\hline
e4
& DSP application branches to an illegal instruction.
& Logic error any place in the code that causes corruption of memory.\\
\hline
e5
& Processing times per crossing are too long.
& SAF reported crossing processing times are consistently falling out of range.\\
\hline
e6
& Too many track segments.  Not necessarily a fault at the source.
& The front-end hardware is malfunctioning; more particles collided than can be managed; bug in the upstream algorithms.\\
\hline
e7
& Corrupt data in a crossing (truncated, misaligned, or bad header).
& Bad checksum or incorrect header data in a crossing due to transmission failure or upstream logic error.\\
\hline
e8
& Corrupt data - no such channels in the detector.
& Logic error in the front-end electronics or firmware (byte swapping).\\
\hline
e9
& Crossing data lost.
& DSP was reset or reboot while an event was being processed; FPGA input queue overflow; FPGA output queue overflow.\\
\hline
e10
& Failed to transfer results down the DSP L1 buffer link (buffer ready flag not set in time).
& The level-1 buffers were not ready to receive data; the farmlet output queues overflowed.\\
\hline
\end{tabular}
\caption{Sample fault scenarios that FVLA is responsible for monitoring.}
\label{fig:faultScenario}
\end{small}
\end{center}
\end{figure*}

Multiple levels of very lightweight agents (VLAs) \cite{jo:lightweightagents03} are one of the primary components responsible for fault mitigation across the BTeV data acquisition system.

The primary objective of the VLA is to provide the BTeV environment with a lightweight, adaptive layer of fault mitigation.  One of the latest phases of work at Syracuse University has involved implementing embedded proactive and reactive rules to handle specific system failure scenarios.  

A scaled prototype of the Level 1 RTES/BTeV environment was presented at the SuperComputing 2003 (SC2003) conference \cite{dm:scproto04}.  Reactive and proactive VLA rules were integrated within this Level 1 prototype and served a primary role in demonstrating the embedded fault tolerant capabilities of the system.

\begin{large}
\vspace*{.23in}
\noindent
\textbf{2.3.  Challenges}
\vspace*{.16in}
\end{large}

While the SC2003 prototype was effective for demonstrating the real-time fault mitigation capabilities of VLAs on limited hardware utilizing 16 DSPs, one of the major challenges is to find out how the behavior of the various levels of VLAs will scale when implemented across the 2500 DSPs projected for BTeV \cite{jk:hardwarefailure03}.  In particular, how frequently should these monitoring tasks be performed to optimize available processing time, and what affect does this have on other components and the overall behavior of a large-scale real-time embedded system such as BTeV.

Given the number of components and countless fault scenarios involved, it is infeasible to design an `expert system' that applies mitigative actions triggered from a central processing unit acting on rules capturing every possible system state.  Instead, the next section describes a distributed approach that uses self-organizing lightweight agents to accomplish fault mitigation within the large-scale real-time RTES/BTeV environement.

\begin{large}
\vspace*{.23in}
\noindent
\textbf{2.4.  SWARM}
\vspace*{.18in}
\end{large}

SWARM (http://www.swarm.org), distributed under the GNU General Public License, is software available as a Java or Objective-C development kit that allows for multi-agent simulation of complex systems \cite{rb:swarmsched97}\cite{md:openframework00}.  It consists of a set of libraries that facilitate implementation of agent-based models.  SWARM has previously been used by the RTES team in simulations that model the RTES/BTeV environment \cite{dm:swarmsubsum04}.

\begin{large}
\vspace*{.17in}
\noindent
\textbf{3.  Self-Organizing VLAs for Real-Time}\\
\hspace*{.13in}\textbf{Scheduling}
\end{large}

\begin{large}
\vspace*{.17in}
\noindent
\textbf{3.1.  Overview}
\end{large}

\hspace*{.13in}

This paper evaluates a \textit{self-organizing} approach that addresses the weaknesses inherent in traditional hierarchical designs.  In this model, rather than hard-wiring the assignment of FVLA role(s) to specific unique DSPs, the DSPs are \textit{polymorphic} in that \textit{every} Worker DSP is equipped to play the role of the FVLA for \textit{any} DSP on the same node.  

The emergent behavior of this design results in self-organization of FVLA responsibilities based on the state and workload of all DSPs within the node at any given point in time.  A certain set of DSPs may play the role of FVLA at one moment, and another set (which may or may not include DSPs from the original set) can be found playing this role later in time.  The organization occurs automatically within the system as performance metrics across DSPs fluctuate.  This eliminates both the financial and efficiency costs associated with having specialized FVLA DSPs that at times sit idle as Worker DSPs operate at full capacity and fall behind on event processing.  It also increases the efficiency of Worker DSPs that may be wasting idle time when crossing processing rates are low.  In effect, a fully connected network of FVLAs is created that will continue to provide effective fault mitigation when exposed to a high volume of system failures.  The key characteristic of this model is that it requires no central management or global processing. 

\begin{large}
\vspace*{.2in}
\noindent
\textbf{3.2.  Cooperative Game Theory\\}
\hspace*{.27in}
\textbf{Scheduling}
\end{large}

\vspace*{.1in}

As outlined above, this approach uses Worker level DSPs to accomplish the tasks that the FVLA is responsible for.  However, these are the same DSPs that are responsible for the critical overall objective of Level 1 physics application (PA) data filtering \cite{sk:pixeldetect02}.  It is therefore extremely important that DSP usage by each Worker VLA is minimal, and only occurs either when the PA is not fully utilizing the DSP, or when emergency fault mitigative action is required. 

Aside from the VLA, there are two additional tasks running on every DSP in RTES at Level 1, namely the Physics Application (PA), and the DSP Kernel/Command Processor itself:\\

\vspace{-.07in}
\noindent
\textbf{Physics Application (PA):}  A typical physics application will read data from the DSP buffer, perform rudimentary checks on data integrity, process data with a specialized physics algorithm, and write results/reports.  The checks include timing, event size, last event time, data integrity, and link failure.  After the data passes the phyics algorithm, the application program checks for logical errors, and for whether or not there have been too many hits to the sensor (too much data).\\

\vspace{-.07in}
\noindent
\textbf{Kernel/Command Processor:}  This provides the basic operating system functionality of the DSP.  Kernel compute cycle consumption should be minimal since it is viewed as overhead from the application's point of view.

As referenced above, game theory has been applied to a wide range of problems, and is used here to coordinate the amount of DSP clock cycle that is allocated between the PA and the VLA.  Both the PA and VLA wish to maximize the number of clock cycles during which they have control.  If the VLA takes too many DSP cycles, then the PA will be unable to process the incoming data at a high enough rate to prevent the buffers from overflowing, resulting in a loss of data continuity.  This is often fatal for the experiment since this lost data could very well contain portions of vital characteristics of the physics properties being evaluated.  If on the other hand, the PA takes too many DSP cycles, then it runs the risk that system faults will go undetected, resulting in acceptance of corrupt data, and/or incremental bottlenecks that again cause buffer overflows.

An efficient adaptive scheduling algorithm is required that will effectively establish scheduling priorities between the PA and VLA.  Mandatory costs associated with the Kernel/Command Processor, including clock cycle costs for context switching must be factored in.  An analysis of the worst-case behavior of tasks (both VLA and PA) can be done to determine the amount of time that must be allotted to each process.  However, there must be a way for the system to adaptively modify these values when environmental conditions change.  That is, if during every interval T, the HEP applications and the operating system use T$_{PA}$ and T$_{OS}$ time units, respectively, then the VLA will be allowed to use T -- T$_{PA}$ -- T$_{OS}$ every T time units \cite{jo:lightweightagents03}.

An analysis of best-case behavior of tasks (VLA and PA) requires the use of a \textit{utility value} in order for each DSP to determine locally precisely when the PA or VLA should relinquish control \cite{ar:gametheory00}.  A reward system based on a combination of the amount of data processed, along with the frequency of VLA maintenance checks, is used by each DSP in calculating the following local utility value :\\

\noindent
DSP Utility Value = Dw$^{-1}$ + cF$^{-1}$ \hspace{.6in} , where\\

\noindent
D = Expected amount of data that DSP could process\\
\hspace*{.25in}during a given time interval (T).

\noindent
w = Current data buffer watermark.

\noindent
F = Total number of clock cycles elapsed since last\\
\hspace*{.25in}FVLA check on neighboring DSPs.

\noindent
c = Adaptive constant representing weight to place on\\
\hspace*{.25in}FVLA checks.\\

Since the amount of data that any single DSP can process over a given time interval (D) is mostly fixed, the utility value essentially involves summing the inverse of the current data buffer watermark (w$^{-1}$) with a weighted value for the inverse of the time elapsed since FVLA functions were last performed (F$^{-1}$).  

The task currently active (PA or VLA) calculates the optimum expected utility value for the DSP every T time units.  If a higher utility value for the DSP is received by remaining active, then the current task will continue.  However, if a higher utility value can be gained by passing control to the currently inactive task (PA or VLA), then that is what it will do.  For example, if the PA is currently active, the input data buffer for a given DSP is low, and FVLA monitoring responsibilities have not been performed on a particular DSP in a long time, then the VLA task will be made active.  If however, the VLA was currently active under these conditions, then the VLA would simply maintain control for another T time steps, at which time corresponding utility values would again be calculated.  This is equivalent to determining :\\

\vspace{-.05in}
max(w, 2 $\times$ ((1 / (1 + e$^{-dF}$)) - .5)\\
\vspace{-.05in}

\noindent
the maximum value of either \textit{w} or 2 $\times$ ((sigmoid function value for \textit{F}) - .5).  Here, 2 $\times$ ((1 / (1 +e$^{-dF}$)) - .5) is an adjusted sigmoid function for F which represent \textit{F} as a weighted value between 0 and 1.  It is important to note here that the value assigned to \textit{d} determines the steepness of the sigmoid function.  In other words, the higher the value of \textit{d}, the higher the adjusted sigmoid value of F.  Rememer that a high value for F means that FVLA tasks are performed more frequently, where as a low value for F means they are performed less often.  The PA is passed (or maintains) control if \textit{w} is higher than this adjusted sigmoid function value for \textit{F}, otherwise the VLA is passed (maintains) control.  For example, if the PA is currently active, the input data buffer watermark for a given DSP is about half full (w=.5), and FVLA functions have recently been performed (the adjusted sigmoid function value for \textit{F} is, say, .15) then the PA will remain active.

\begin{large}
\vspace*{.2in}
\noindent
\textbf{4.  Results}\\
\vspace*{-.1in}
\end{large}

SWARM simulates Farmlet data buffer queues that are populated at a rate consistent with the behavior of the incoming physics crossing data.  Each DSP within a given Farmlet processes a fixed amount of data at each discrete time step.  Errors are introduced randomly within each Worker DSP at a fixed rate using a Multiply With Carry (RWC8gen) random number generator with a fixed seed.  Any time a software or hardware error is encountered within the simulation, the processing rate for that DSP decreases a set amount depending on the type of  error.  The error is cleared when any DSP within the same Farmlet performs FVLA checks against the DSP with the error.  However, there is a time cost associated with performing these checks.  As detailed in the section above describing the self-organizing model, the DSP must decide whether or not it is worth taking time to perform FVLA monitoring tasks against neighboring DSPs.  If checks are performed too frequently, then the time available for data crossing processing is limited.  On the other hand, if they are not performed frequently enough, then the chances that other DSPs within the same Farmlet are experiencing errors is high.  As described, a high error rate will also lead to slow processing rates.  

The formula designed for these experiments calculates the frequency of performing FVLA tasks for neighboring DSPs as a sigmoid function adjusted to a value between 0.0 and 1.0.  This is compared against the watermark for the crossing data buffer, and the DSP makes a decision on where to devote its energy, as described in detail in the last section.

The decision of whether the VLA or PA has control of the DSP is made by each DSP at each and every time step in the SWARM simulation.  In this way, the monitoring tasks required by the environment are always met, but not necessarily by one (or a few) designated DSPs.  Instead, these tasks are performed by any \textit{polymorphic} DSP within the Farmlet as dictated by the changing needs of the environment.  The DSPs themselves \textit{self-organize} as different DSPs within the Farmlet take on the necessary monitoring tasks at different points in time as required by the environment.

Multiple sets of experiments were run using 12 distinct d-values for the sigmoid function ranging from .0001 to 3.0.  This was repeated for each of 5 distinct error rates ranging from .00001 to .1.  The fixed error rate represents the probability of an error occurring at any given node during a single time step.  For each unique error rate and d-value, the average number of crossings processed over a fixed time period (in this case 10,000 SWARM simulation time steps) was recorded to measure data throughput.

\begin{figure}[t]
\includegraphics[width=103mm, height=90mm]{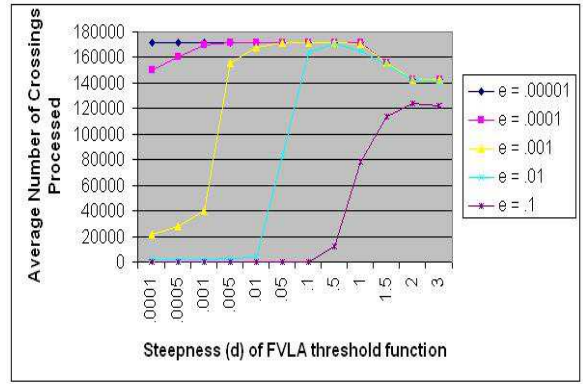}
\vspace*{-1.725in}
\caption{The average number of crossings processed at various fixed error rates vs. (d) values representing the steepness of the FVLA threshold sigmoid function.}
\label{fig:totald}
\end{figure}

\begin{figure}[t]
\centering
\includegraphics[width=126mm]{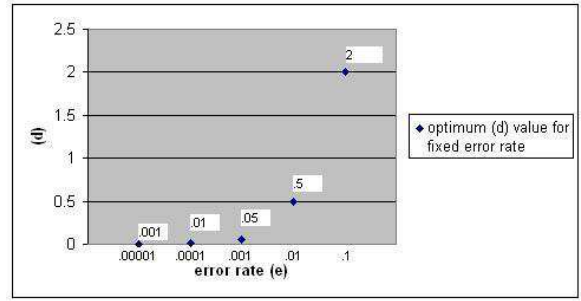}
\vspace*{-2.15in}
\caption{The optimum (d) value found experimentally across fixed error rates (e).}
\label{fig:optimumd}
\end{figure}

The results of these experiments are shown as graphs in Figures \ref{fig:totald} and \ref{fig:optimumd} which demonstrate that an optimum d-value can be found experimentally for any given error rate (e).  For example, Figure \ref{fig:totald} shows that for a fixed error rate of .1, the optimum d-value was found to be approximately 2.0, at which point 125000 crossings were processed.  Figure \ref{fig:optimumd} shows the optimum d-value found for each distinct fixed error rate.  For example, the optimum d-value found is .01 given a fixed error rate (e) of .0001, is .05 given an error rate of .001, is .01 given a fixed error rate of .5, and so on.  Clearly, the total amount of data processed by the system continues to decrease as the frequency of FVLA tasks being performed continues to drop (the d-value decreases) below the experimental optimum d-value threshold.  This was expected since this essentially means that software and/or hardware faults are occurring at a faster rate than they are being monitored and corrected by the FVLA, resulting in the loss of particular software/hardware components that could have contributed to a higher crossing processing rate.  Similarly, as the frequency of FVLA tasks being performed increases (the d-value increases) past the optimum value, the DSP is spending unnecessary excess time performing FVLA monitoring tasks.  Since this is time that it could have spent instead performing crossing processing, the data crossing processing rate drops.

Another finding demonstrated by these results is that it is far more detrimental to not perform FVLA monitoring tasks frequently enough, as compared to performing them too often.  Figure \ref{fig:totald} shows the comparatively minimal cost of exceeding the optimum d-value, as opposed to the high cost of it being too low.  This confirms initial intuitions that the cost of individual errors occurring too frequently far outweighs the costs associated with performing individual FVLA monitoring tasks.

The optimum d-values experimentally found for each fixed error rate are shown in Figure \ref{fig:optimumd}.  These values demonstrate another expected trend in the experiments, namely that the optimum frequency of FVLA monitoring tasks increases as the error rate increases.  In other words, as more faults occur across the system, more FVLA monitoring and mitigation tasks must be performed.

One of the most valuable results of these experiments for the RTES collaboration is that it has demonstrated an experimental method for finding an optimum d-value given a fixed error rate.

\begin{large}
\vspace*{.18in}
\noindent
\textbf{5.  Next Steps}\\
\vspace*{-.04in}
\end{large}

The results presented in this paper have provided a way to experimentally determine optimum systems scheduling thresholds using polymorphic, lightweight, self-organizing agents within large-scale, real-time systems.  However, error rates, particularly within these types of real-time systems are typically not fixed.  Therefore, the next step for the RTES collaboration is to extend the adaptability of the agents to handle fluctuations in the error rate across components of the system.  Variations of reinforcement learning and other evolutionary techniques are currently being evaluated for ways that they may be used to automatically adapt to changing system states.  In particular, alternative reward distribution techniques may be necessary to best handle the scale, complexity, and uncertainty inherent in fault mitigation for large-scale systems such as BTeV.

In addition, another scaled prototype of the actual projected RTES/BTeV software and hardware environment based on the SC2003 demonstration system is also currently being developed, and will integrate the VLA self-organizing model.

\begin{large}
\vspace*{.2in}
\noindent
\textbf{6.  Conclusion}\\
\vspace*{-.04in}
\end{large}

This paper has described a self-organizing, multi-agent systems approach for systems scheduling based on cooperative game theory.  The design uses distributed, lightweight agents embedded within individual DSPs to perform Farmlet level monitoring tasks.  The results show that the method is able to experimentally determine the optimum rate at which certain tasks should be performed at each DSP based on specific error rates.  The experiments were run using a SWARM simulation of2500 DSPs from a data acquisition system for a High Energy Physics particle accelerator being developed at Fermi National Accelerator Laboratory.  Future work will focus on evolving these rates dynamically based on fluctuating error rates experienced within the system.

\begin{large}\textbf{\\
Acknowledgements}\\
\end{large}
\vspace{-.05in}
\\
The research conducted was sponsored by the National Science Foundation in conjunction with Fermi National Laboratories, under the BTeV Project, and in association with RTES, the Real-time, Embedded Systems Group.  This work has been performed under NSF grant \# ACI-0121658.

\vspace{-.07in}
\bibliography{his04}

\end{document}